# A Domain-Independent Algorithm for Plan Adaptation

**Steve Hanks**                                HANKS@CS.WASHINGTON.EDU
**Daniel S. Weld**                             WELD@CS.WASHINGTON.EDU
*Department of Computer Science and Engineering*
*University of Washington*
*Seattle, WA 98195*

## Abstract

The paradigms of transformational planning, case-based planning, and plan debugging all involve a process known as *plan adaptation* — modifying or repairing an old plan so it solves a new problem. In this paper we provide a domain-independent algorithm for plan adaptation, demonstrate that it is sound, complete, and systematic, and compare it to other adaptation algorithms in the literature.

Our approach is based on a view of planning as searching a graph of partial plans. Generative planning starts at the graph's root and moves from node to node using plan-refinement operators. In planning by adaptation, a library plan—an arbitrary node in the plan graph—is the starting point for the search, and the plan-adaptation algorithm can apply both the same refinement operators available to a generative planner and can also retract constraints and steps from the plan. Our algorithm's completeness ensures that the adaptation algorithm will eventually search the entire graph and its systematicity ensures that it will do so without redundantly searching any parts of the graph.

## 1. Introduction

Planning by adapting previously successful plans is an attractive reasoning paradigm for several reasons. First, cognitive studies suggest that human experts depend on a knowledge of past problems and solutions for good problem-solving performance. Second, computational complexity arguments show that reasoning from first principles requires time exponential in the size of the problem. Systems that reuse old solutions can potentially avoid this problem by solving a smaller problem: that of adapting a previous solution to the current task. Intuition tells us that many new problem-solving situations closely resemble old situations, therefore there may be advantage to using past successes to solve new problems.

For example, case-based planners typically accomplish their task in three phases:

- **RETRIEVAL**: Given a set of initial conditions and goals, retrieve from the library a similar plan—one that has worked in circumstances that resemble the inputs. The retrieval phase may also involve some superficial modification of the library plan, for example, renaming constants and making the library plan's initial and goal conditions match the input specifications.

- **ADAPTATION**: Modify the retrieved plan — e.g., by adding and removing steps, by changing step orders, or by modfying variable bindings — until the resulting plan achieves the input goal.

- **GENERALIZATION**: Generalize the newly created plan and store it as a new case in the library (provided it is sufficiently different from plans currently in the library).

This paper focuses on the adaptation process in the general context of a case-based planning system; however, our adaptation algorithm could be useful for transformational and plan debugging systems as well.

## 1.1 Motivation

Work in case-based planning has historically been conducted in particular application domains, and has tended to focus on representation rather than algorithmic issues. The research addresses problems like what features of a library plan make good indices for subsequent retrieval, how features of the library plan can suggest effective adaptation strategies, and so on.

Our work develops a domain-independent algorithm for plan adaptation, and is therefore complementary: it provides a common platform with which one can analyze and compare the various representation schemes and adaptation strategies, as well as explore in the abstract the potential benefits of the case-based approach to planning. Sections 7 and 8.2 discuss the CHEF (Hammond, 1989) and PRIAR (Kambhampati & Hendler, 1992) systems using our framework, and Section 6.1 characterizes precisely the potential benefits of plan adaptation versus plan generation.

This paper presents an algorithm, SPA (the "systematic plan adaptor") for plan adaptation that is sound, complete, and systematic. Soundness means that the output plan is guaranteed to satisfy the goal, completeness means that the planner will always find a solution plan if one exists (regardless of the library plan provided by the retriever), and systematicity means that the algorithm explores the space of adaptations non-redundantly (in short, it will never consider an adaptation more than once).

Systematicity is the trickiest property to guarantee, and for two reasons. First, the adapter operates in a space of incomplete plans.[1] Each incomplete plan can expand into an exponential number of completions; systematicity requires that the adaptation algorithm never consider two incomplete plans that share even one completion, whereas completeness requires that every potential completion be considered. Second, plan adaptation requires a combination of retracting previous planning decisions (choice and ordering of plan steps, binding of variables within the action schemas), as well as making new decisions. Systematicity requires that a decision, once retracted, never be considered again.

Our framework for planning by adaptation is based on two premises having to do with the nature of stored plans and how they are manipulated:

- A library plan or case is stored as a complete and consistent plan for solving the prior problem. This plan contains the steps and orderings that solved the prior problem along with additional constraints and dependency information that record why the steps and orderings appear there. Applying a case to a new problem first involves adjusting the library plan to match the initial and goal conditions of the current problem, a process that produces a consistent but incomplete plan. The adaptation process attempts to complete this plan.

---

1. An incomplete plan may be partially ordered, may contain partially constrained variables, and may require additional steps or constraints for it to achieve the goal.

- The adaptation process consists of a standard set of plan-refinement operators (those that add steps and constraints to the plan) plus the ability to retract the refinements made when the library plan was originally generated.

We view the general planning problem as a search through a directed graph of partial plans. The graph's root represents the null plan and its leaves represent finished plans. Generative planning starts at the root of the graph and searches for a node (plan) that satisfies the goal. It generates the graph by successively refining (constraining) the plan. The retrieval phase of an adaptation-based planner, on the other hand, returns an arbitrary node in the graph, and the adapter begins searching from that point. It must be able to search down the graph like a generative planner but also must be able to search backward through the graph by retracting constraints, producing more abstract plans. Our complete and systematic adaptation algorithm is able to search every node in the graph without considering any node more than once.

We have implemented our algorithm[2] in Common Lisp on UNIX workstations and tested it on several problem domains. Experimental studies compare our algorithm to a similar effort, PRIAR (Kambhampati & Hendler, 1992). Our results show a systematic speedup from plan reuse for certain simple and regular problem classes.

Our work on SPA makes the following contributions:

- Our algorithm captures the essence of the plan-adaptation process within an extremely simple framework. As such it is amenable to formal analysis, and provides a framework with which to evaluate other domain-independent algorithms like the PRIAR system (Kambhampati & Hendler, 1992), and to analyze domain-dependent representations and adaptation strategies.

- We use the framework to investigate CHEF's transformational approach to plan adaptation, and show how CHEF's repair strategies could be added to SPA as search-control heuristics.

- We analyze the tradeoff between plan generation and adaptation, characterizing the similarity required by the plan retrieval routine to produce speedup.

- We report on empirical experiments and demonstrate for a simple class of problems a systematic relationship between computation time and the similarity between the input problem and the library plan the adaptation algorithm begins with.

The paper proceeds as follows: we first review previous work in planning by adapting or repairing previous solutions. Next we review the least-commitment generative planning algorithm on which SPA is based, in doing so introducing many of SPA's data structures. Section 4 then explains the details of our adaptation algorithm. In Section 5 we prove that SPA is sound, complete and systematic.

Since the speed of adaptation depends on the quality of the plan retrieved from the library, it can be faster to perform generative planning than attempt to adapt an inappropriate library plan; in Section 6 we analyze this tradeoff and also discuss some interesting interactions between the algorithms for adaptation and plan fitting. Then in Section 7 we show how transformational planners such as Hammond's (1990) CHEF system can be analyzed using our framework. Section 8 reports our empirical studies. After reviewing related work (Section 9), Section 10 discusses our progress and poses questions for future research.

---

2. Send mail to bug-spa@cs.washington.edu for information on acquiring free source code via FTP.

## 2. Previous Work on Adaptive Planning

The idea of planning by adaptation has been in the literature for many years, and in many different forms. In this section we review this work briefly, trying to motivate and put into perspective our current work on SPA.

The basic idea behind planning by adaptation (or similar work in case-based planning, transformational planning, or planning by solution replay) is to solve a new problem by (1) retrieving from memory a problem that had been solved previously, then (2) adapting the old solution to the new problem.

The CHEF system (Hammond, 1990) is a case-based planner that solves problems in the domain of Szechwan cooking. When given a goal to produce a dish with particular properties, CHEF first tries to *anticipate* any problems or conflicts that might arise from the new goal, and uses that analysis to *retrieve* from memory a candidate solution (baseline plan). The baseline plan is then manipulated by a *modifying* algorithm that tries to satisfy any new goals and repair problems that did not arise in the baseline scenario. It then executes the plan, and if execution results in failure a *repair* algorithm analyzes the failure and uses the result of that analysis to improve the index for this solution so that it will not be retrieved in situations where it will fail again.

CHEF addresses a wide range of problems important to case-based planning: how to anticipate problems, how to retrieve a solution from the case library, how to adapt or modify an old solution, and how to use execution failure to improve subsequent retrievals. Our SPA system primarily addresses the adaptation problem, and in Section 7 we use our framework to analyze CHEF's modification strategies in some detail.

The PLEXUS system (Alterman, 1988) confronts the problem of "adaptive planning," but also addresses the problem of *run-time* adaptation to plan failure. PLEXUS approaches plan adaptation with a combination of tactical control and situation matching. When a plan failure is detected it is classified as being either a failing precondition, a failing outcome, a case of differing goals, or a step out of order. Ignoring the aspects of PLEXUS that deal with incomplete and incorrect knowledge, the program's main repair strategy involves replacing a failed plan step with one that might achieve the same purpose. PLEXUS uses a semantic network to represent abstraction classes of actions that achieve the same purpose (walking and driving are both instances of transportation actions, for example).

The GORDIUS system (Simmons, 1988) is a *transformational planner*. While the difference between a transformational planner and a case-based planner has not been precisely defined, a major difference concerns how the two types of planners get the starting point for plan adaptation. Cased-based systems get this plan via retrieval of a past solution from a library, but GORDIUS combines small plan fragments for different (hopefully independent) aspects of the current problem. GORDIUS differs from CHEF in two other ways: first of all, GORDIUS does not perform an *anticipation* analysis on the plan, trying to identify trouble spots before library retrieval. Instead it accepts the fact that the retrieved plan will be flawed, and counts on its repair heuristics to patch it. CHEF, on the other hand, assumes that the retrieved library plan will be a close enough fit to the new problem so that little or no adaptation will be necessary. Second, much of the GORDIUS work is devoted to developing a set of repair operators for quantified and metric variables.

The main idea behind the SPA system separates it from the three systems mentioned above: that the process of plan *adaptation* is a fairly simple extension to the process of plan *generation*. As a consequence we can assume that the algorithm that generates library plans—and the structure of those plans—is the same as the adaptation algorithm and the plan structures it generates. In the SPA view, plan generation is just a special case of plan adaptation (one in which there is no retrieved structure to exploit).

Two pieces of work developed at the same time as SPA adopt similar assumptions: the PRIAR system (Kambhampati & Hendler, 1992) and the NoLIMIT system (Veloso, 1992, 1994).

The main difference between SPA and PRIAR is the underlying planning algorithm: SPA uses a constraint-posting technique similar to Chapman's (1987) TWEAK as modified by McAllester and Rosenblitt (1991), whereas PRIAR uses a variant of NONLIN (Tate, 1977), a hierarchical planner. Section 8 compares these two planners in some detail.

The NoLIMIT system also takes a search-oriented approach to planning. It differs from SPA in the role a case plays in the problem-solving process. A library plan (case) in a transformational or case-based-planning framework stores a solution to a prior problem along with a summary of what new problems it would be a suitable solution for, but it contains little information about the *process* that generated the solution. Derivational analogy, on the other hand, stores substantial descriptions of the *decisions* that resulted in the solution. In particular, Veloso's system records more information at each choice point than does SPA: a list of failed alternatives, for example. The relative effectiveness of the two approaches seems to hinge on the extent to which old planning decisions (as opposed to the plans themselves) can be understood and exploited in similar planning episodes.

In summary, we consider our work on SPA to be complementary to most existing work in transformational or case-based planning. The latter has concentrated on developing heuristically effective problem solvers for particular domains. Case-based-planning research has also explored the problem of how to retrieve cases from the plan library—in particular the problem of how to index them effectively. SPA, on the other hand, is a domain-independent algorithm, and does not address the retrieval or indexing problems in any deep way.

The main objectives of this work are (1) to explore the idea that plan adaptation is a fairly minor representational and algorithmic variant of the basic problem of plan generation, (2) to provide preliminary evidence that this view of plan adaptation is empirically viable, and (3) to provide to the community an implementation of an algorithm that will allow effective problem solvers to be built based on this idea.

We now begin the development of our framework with a description of the underlying framework for purely generative planning.

## 3. Generative Planning: the SNLP Algorithm

Since the SPA algorithm is an extension of a partial-order, constraint-posting, least commitment generative planning algorithm, we begin by presenting the generation algorithm itself. However, we do so using the notation of the SPA system, and in the process intro-duce many of the data structures and functions needed to implement the full adaptation algorithm. Our treatment is brief—see elsewhere (McAllester & Rosenblitt, 1991; Barrett & Weld, 1994a) for more detail.

## 3.1 Data Structures

An *action* is a schematic representation of an operator available to the planner. An action consists of a *name*, a set of *preconditions*, an *add list*, a *delete list*, and a set of *binding constraints*. The first four are *expressions* that can contain *variables*. We use question marks to identify variables, `?x` for instance. Binding constraints are used to indicate that a particular variable cannot be bound to a particular constant or to some other variable. Here is an action corresponding to a simple blocksworld `puton` operator:

```
(defaction :name        '(puton ?x ?y)
           :preconds    '((on ?x ?z) (clear ?x) (clear ?y))
           :adds        '((on ?x ?y) (clear ?z))
           :deletes     '((on ?x ?z) (clear ?y))
           :constraints '((<> ?x ?y) (<> ?x ?z) (<> ?y ?z)
                          (<> ?x TABLE) (<> ?y TABLE)))
```

An instance of an action is inserted into a plan as a *step*. Instantiating an action involves (1) giving unique names to the variables in the action, and (2) assigning the step a unique index in the plan, so a plan can contain more than one instance of the same action. A step is therefore an instance of an action inserted into a plan with an index that uniquely identifies it.

A plan also contains a set of *constraints*, which either constrain the order of two steps in the plan or constrain the bindings of variables in the steps. An ordering constraint takes the form $S_i < S_j$, where $S_i$ and $S_j$ are steps, and indicates that the step with index $i$ must occur before the step with index $j$. A binding constraint is of the form $(= v_1 v_2)$ or $(\neq v_1 v_2)$, where $v_1$ is a variable appearing in some step in the plan and $v_2$ is either a variable or constant appearing in the plan.[3]

We also annotate every constraint with a record of why it was placed in the plan. Therefore a plan's constraints is actually a set of pairs of the form $\prec c, r \succ$ where $c$ is a either an ordering or binding constraint, and $r$ is a *reason* data structure (defined below).

The final component of a plan is a set of causal *links*, each of the form $S_i \xrightarrow{Q} S_j$, where $S_i$ and $S_j$ are steps, and $Q$ is an expression. The link records the fact that one purpose of $S_i$ in the plan is to make $Q$ true, where $Q$ is a precondition of $S_j$. If a plan contains a link $S_i \xrightarrow{Q} S_j$ it must also contain the ordering $S_i < S_j$.

A plan consists of a set of steps, a set of constraints, and a set of links.

A *planning problem* is a triple $\prec \mathcal{I}, \mathcal{G}, \texttt{Actions} \succ$. $\mathcal{I}$ is a set of expressions describing the problem's initial conditions, $\mathcal{G}$ is a set of expressions describing the problem's goal, and `Actions` is the set of available actions. We assume that `Actions` is available to the algorithm as a global variable.

---

3. The `:constraints` field in an action description also contains binding constraints: `(<> ?x ?y)` is equivalent to the $(\neq$ ?x?y$)$ notation used in the rest of the paper.

Next we exploit a standard representational trick and convert a planning problem to a plan by building a plan that contains

1. a step with name *initial*, index 0, having no preconditions nor delete list, but with an add list consisting of the problem's initial conditions $\mathcal{I}$,

2. a step with name *goal*, index $\infty$, with preconditions consisting of the goal expressions $\mathcal{G}$, but empty add and delete lists,

3. the single ordering constraint $S_0 < S_\infty$,

4. no variable-binding constraints,

5. no links.

Every plan must contain at least the two steps and the ordering, and we call the plan with only this structure the *null plan*.

Two interesting properties of a plan are its set of *open preconditions* and its set of *threatened links*. The former is the set of expressions that appear in any step's precondition but have no causal support within the plan; the latter is the set of explicit causal relationships that might be nullified by other steps in the plan. Formally an open condition in a plan, notated $\xrightarrow{Q} S_j$, is a step $S_j$ in the plan that has precondition $Q$, and for which there is no link in the plan of the form $S_i \xrightarrow{Q} S_j$ for any step $S_i$.

A link of the form $S_i \xrightarrow{Q} S_j$ is *threatened* just in case there is another step $S_t$ in the plan such that

1. the plan's ordering constraints would allow $S_t$ to be ordered after $S_i$ and before $S_j$, and

2. $S_t$ has a postcondition (either add[4] or delete) that the plan's variable-binding constraints would allow to unify with $Q$.

A plan with no open preconditions and no threatened links is called a *solution* to the associated planning problem.

Finally we introduce the *reason* data structure, unnecessary for generative planning but essential for adaptation. Every time a step, link or constraint is added to a plan an associated *reason* records its purpose. A reason consists of two parts: 1) a symbol recording why the constraint was added (either ADD-STEP, ESTABLISH, or PROTECT), and 2) either a link, step, or threat in the plan identifying the part of the plan being repaired. Section 3.4 discusses reasons in more detail.

---

4. Some people find it counterintuitive that $S_t$ should threaten $S_i \xrightarrow{Q} S_j$ if it has $Q$ on its *add* list. After all, the presence of $S_t$ doesn't prevent $Q$ from being true when $S_j$ is executed. Our definition, adopted from McAllester and Rosenblitt (1991), is necessary to ensure systematicity. See (Kambhampati, 1993) for a discussion.

### 3.2 The Planning Algorithm

The generative planning algorithm is based on the idea of starting with the null plan and successively *refining* it by choosing a flaw (open condition or threatened link) and adding new steps, links, or constraints to fix it. The algorithm terminates either when a complete plan is found (success) or when all possible refinement options have been exhausted (failure).

Consider a planning problem with initial conditions `Initial` and goal `Goal`. We assume throughout the paper that the set of actions available to the planner, `Actions`, is fixed. We now define a top-level function, `PlanGeneratively`, which initializes the search and calls a function that performs the refinement process.

**function PlanGeneratively(`Initial`, `Goal`)**: Plan or *failure*
1    `N` := Construct the null plan from `Initial` and `Goal`
2    **return** `RefinementLoop(N)`

`RefinementLoop` searches through the space of partial plans for a solution plan, storing the search horizon internally, each time choosing a plan and calling `RefinePlan`, which chooses and repairs a single flaw in that plan.

**function RefinementLoop(`NullPlan`)**: Plan or *failure*
1      Frontier := {`NullPlan`}
2      **loop forever:**
3          **if** Frontier is empty **then return** *failure*
4          P := select an element from Frontier
5          Delete P from Frontier
6          **if** P is a solution **then return** P
7             **else** add all elements of `RefinePlan(P)` to Frontier

Refining a plan consists of two parts: selecting a flaw in the plan (an open precondition or threatened link), then generating all possible corrections to the flaw. The selection of which flaw to correct need not be reconsidered, but the manner in which it is corrected might have to be, which is why all possible corrections are added to the search frontier.

**function RefinePlan(`P`)**: List of plans
1    F := Select a flaw from P
2    **return** `CorrectFlaw(F, P)`

Correcting a flaw amounts to resolving an open condition or resolving a threat:

**function CorrectFlaw(`F`, `P`)**: List of plans
1    **if** F is an open precondition **then**
2        **return** `ResolveOpen(F, P)`
3        **else return** `ResolveThreat(F, P)`

An open condition can be supported either by choosing an existing step that asserts the proposition or by adding a new step that does so:

**function ResolveOpen($\xrightarrow{Q}$S$_j$, P): List of plans**

1     **for** each step S$_i$ currently in P **do**
2        **if** S$_i$ can be ordered before S$_j$, and S$_i$ adds a condition unifying with Q **then**
3           **collect** Support(S$_i$, Q, S$_j$, P)
4     **for** each action A in **Actions** whose add list contains a condition unifying with Q **do**
5        (S$_k$, P') := AddStep(A,P)
6        **collect** Support(S$_k$, Q, S$_j$, P')
7     **return** the list of plans collected at lines 3 and 6.

The function **AddStep** takes an action and plan as inputs, makes a copy of the plan, instantiates the action into a step, and adds it to the plan with the required ordering and binding constraints. It returns both the newly added step and the newly copied plan.

**function AddStep(A, P): (Step, Plan)**

1     S$_k$ := a new step with action A and an index unique to P
2     R := a new reason of the form [ADD-STEP S$_k$]
3     P' := a copy of P
4     Add S$_k$ to P'
5     Add each of A's :constraints to P', each tagged with R
6     Add the orderings S$_0$ < S$_k$ and S$_k$ < S$_\infty$ to P', both tagged with R
7     **return** (S$_k$, P')

Now **Support** adds a causal link between two existing steps in the plan S$_i$ and S$_j$, along with the required ordering and binding constraints. Notice that there might be more than one way to link the two steps because there might be more than one postcondition of S$_i$ that can unify with the link proposition Q. This operation is identical to the way SNLP adds causal links except that the constraints are annotated with a *reason* structure.

**function Support(S$_i$, Q, S$_j$, P): List of plans**

1     **for** each set of bindings B causing S$_i$ to assert Q **do**
2        P' := a copy of P
3        L := a new link S$_i$$\xrightarrow{Q}$S$_j$
4        R := a new reason [ESTABLISH L]
5        Add L to P'
6        Add the ordering constraint S$_i$ < S$_j$ to P', tagged with R
7        Add B to P', tagged with R
8        **collect** P'
9     **return** the set of plans collected at step 8





Recall that a threat to a link $S_i \xrightarrow{Q} S_j$ is a step $S_t$ that can consistently be ordered between $S_i$ and $S_j$ and can consistently assert either $Q$ or $\neg Q$ as a postcondition (i.e. either adds or deletes $Q$). We use the notation $\prec S_i \xrightarrow{Q} S_j, S_t \succ$ to denote this threat. The three possible ways to resolve a threat—promotion, demotion, and separation—involve adding ordering and binding constraints to the plan:

**function ResolveThreat($\prec S_i \xrightarrow{Q} S_j, S_t \succ$, P):** List of plans

1   R := a new reason [PROTECT $\prec S_i \xrightarrow{Q} S_j, S_t \succ$]
2   **if** $S_t$ can be consistently ordered before $S_i$ **then**
3       P′ := a copy of P
4       Add the constraint $S_t < S_i$ to P′, tagged with R
5       **collect** P′
6   **if** $S_t$ can consistently be ordered after $S_j$ **then**
7       P′ := a copy of P
8       Add the constraint $S_j < S_t$ to P′, tagged with R
9       **collect** P′
10  **for** each set of bindings B that prevents $S_t$'s effects from unifying with Q **do**
11      P′ := a copy of P
12      Add constraints $S_i < S_t$ and $S_t < S_j$ to P′, both tagged with R
13      Add B to P′, tagged with R
14      **collect** P′
15  **return** all new plans collected at lines 5, 9, and 14 above

Note that line 10 is a bit subtle because both codesignation and noncodesignation constraints must be added.[5] For example, there are two different minimal sets of binding constraints that must be added to protect $S_i \xrightarrow{(\text{on ?x ?y})} S_j$ from a step $S_t$ that deletes (on ?a ?b): $\{(\neq ?x ?a)\}$, and $\{(= ?x ?a), (\neq ?y ?b)\}$. Line 12 is also interesting — the constraints $S_i < S_t$ and $S_t < S_j$ are added in order to assure systematicity.

### 3.3 Formal Properties: Soundness, Completeness, Systematicity

McAllester and Rosenblitt (1991) prove three properties of this algorithm:

1. Soundness: for any input problem with initial conditions I, goal G, and actions Actions, if PlanGener#atively(I, G) successfully returns a plan $\mathcal{P}$, then executing the steps in $\mathcal{P}$ in any situation satisfying I will always produce a state in which G is true.

2. Completeness: PlanGeneratively will find a solution plan if one exists.

3. Systematicity: PlanGeneratively will never consider the same plan (partial or complete) more than once.

Completeness and systematicity can be explained further by viewing PlanGeneratively as searching a directed graph of partial plans. The graph has a unique root, the null plan,

---

5. But see (Peot & Smith, 1993) for an alternative approach.





and a call to `RefinePlan` generates a node's children by choosing a flaw and generating its successors (the partial plans resulting from considering all possible ways of fixing it). Figure 2 (Page 330) illustrates how refinement replaces a frontier node with its children.

The completeness result means simply that every solution plan appears as a leaf node in this graph, and that **PlanGeneratively** will eventually visit every leaf node if necessary.

Systematicity implies that this directed graph is in fact a *tree*, as Figure 2 suggests. The graph has this property because the children of a partial plan node are all alternative fixes for a flaw *f*—each child has a different step, ordering, or binding constraint added to fix *f*. And since subsequent refinements only add more constraints, each of *its* children inherit this commitment to how *f* should be fixed. Therefore any plan on the frontier will differ from every other plan on the frontier in the way it fixes *some* flaw, and the same plan will never appear on the frontier more than once.

### 3.4 Using *Reasons* to Record Refinement Decisions

As we mentioned above, the reason data structure is unnecessary in a planner that performs only refinement operations. SNLP, for example, does not use them. However, they provide the basis for retracting past decisions which is a necessary component of plan adaptation as discussed in the next section. Before explaining the retraction process, however, we summarize the *reason* data structures that record how and why a plan was refined. A different reason structure is used for each of the three types of refinement:

- Step addition. When a new step $S_i$ is added to a plan (function `AddStep`), the variable-binding constraints associated with its action schema are also added, along with two ordering constraints ensuring that the new step occurs after the initial step and before the goal step. The reasons accompanying these constraints are all of the form [ADD-STEP $S_i$].

- Causal link addition. When a link of the form $S_i \xrightarrow{Q} S_j$ is added to the plan (function `Support`), an ordering constraint $S_i < S_j$ is also added, along with variable-binding constraints ensuring that the selected postcondition of $S_i$ actually asserts the proposition required by the selected precondition of $S_j$. These constraints will be annotated with a reason structure of the form [ESTABLISH $S_i \xrightarrow{Q} S_j$].

- Threat resolution. When a link $S_i \xrightarrow{Q} S_j$ is threatened by a step $S_t$, the link can be resolved (function `ResolveThreat`) by adding one of three sorts of constraints: an ordering of the form $S_t < S_i$, an ordering of the form $S_j < S_t$, or variable-binding constraints ensuring that the threatening postcondition of $S_t$ does not actually falsify the link's proposition $Q$. These constraints will be annotated with a reason structure of the form [PROTECT $\prec S_i \xrightarrow{Q} S_j, S_t \succ$].

This completes our review of generative (refinement) planning, so we now turn to the extensions that turn this planner into an adaptive algorithm.

## 4. Plan Adaptation

There are two major differences between generative and adaptive planning:





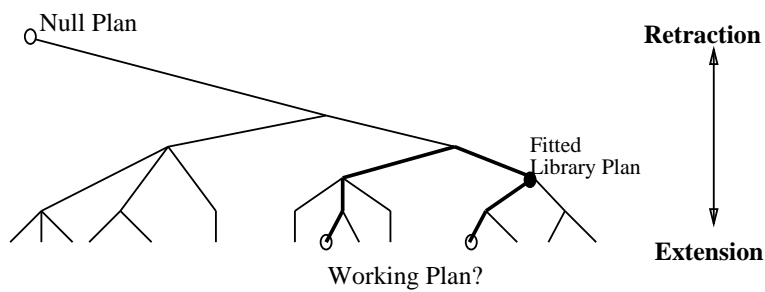

Figure 1: Plan refinement and retraction as search in **plan** space.

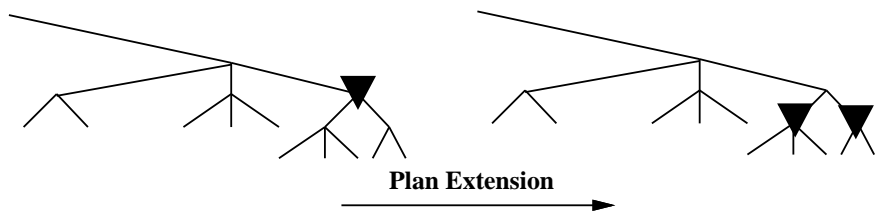

Figure 2: Plan refinement replaces a plan tagged DOWN with a set of new plans, each with an additional step or constraint.

1. In adaptive planning there is a *library retrieval* phase in which the plan library is searched for a plan that closely matches the input initial and goal forms; the library plan is then *adjusted* to match the current problem.

2. Adaptive planning begins with this retrieved and adjusted partial plan, and can retract planning constraints added when the plan was originally generated; generative planning begins with a plan with no constraints and can only add new ones.

In other words, both generative and adaptive planning are searching for a solution plan in a tree of partial plans, but generative planning starts at the (unique) root of the plan tree whereas adaptive planning begins at some arbitrary place in the tree (possibly at a solution, possibly at the root, possibly at some interior node).

Figure 1 shows that adaptation starts at an interior node, and a solution might appear "below" it in the tree or in a different subtree altogether. As a result the adaptation algorithm must be able to move "up" the tree by *removing* constraints from the plan as well as move "down" the tree by *adding* constraints.

Figures 2 and 3 show the way this movement is accomplished: plan *refinement* is the (only) operation performed by a generative planner. It takes a partial plan on the horizon and replaces it with that plan's *children*, a set of plans identical to the input plan except for having one more flaw repaired.

Plan *retraction* takes a plan on the horizon and chooses a causal link, or set of constraints to *remove*. That plan is replaced on the horizon with the parent (which is marked for additional retraction), along with the plan's siblings (representing all *alternative* ways of





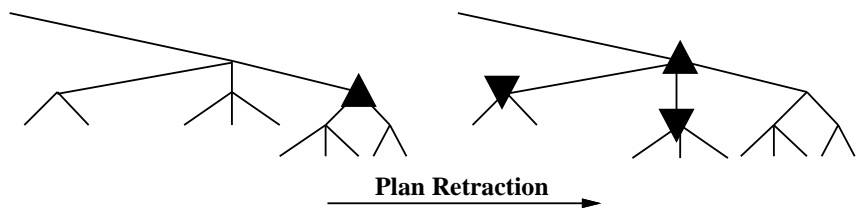

**Plan Retraction**

Figure 3: Plan retraction replaces an UP tagged plan with another UP plan and several sibling plans tagged DOWN.

re-fixing the flaw whose fix was retracted from the plan). The siblings are then tagged for additional refinement.

In Section 5 we show that this simple scheme—augmenting a generative planner's refinement ability with the ability to retract constraints—is sufficient to implement plan adaptation. In other words, we prove that the adaptive planner will still only produce valid solutions (soundness), it can find a solution no matter where in the plan space the library plan places it initially (completeness), and it still doesn't explore areas of the plan tree redundantly (systematicity).

## 4.1 The Adaptive Planning Algorithm

The adaptation algorithm performs a standard breadth-first search, maintaining the search frontier as a set of pairs each of the form $\prec\mathcal{P}$, UP$\succ$ or form $\prec\mathcal{P}$, DOWN$\succ$. In either case $\mathcal{P}$ is a (possibly incomplete) plan and UP or DOWN indicates the way to manipulate the plan to generate the plan's neighbors in the search space: DOWN means generate $\mathcal{P}$'s successors by further refining it (adding new steps and/or constraints) exactly as in generative planning; UP means generate $\mathcal{P}$'s successors by retracting one of the refinements made when the plan was originally constructed.

    **function PlanAdaptively(Initial, Goal, Library)**: Plan or *failure*

1      LibPlan := *retrieve* a plan for Initial and Goal from Library
2      AdjustedPlan := *adjust* LibPlan to match Initial and Goal exactly
3      NewPlan := AdaptationLoop(LibPlan)
4      *Store* NewPlan in Library
5      **return** NewPlan

### 4.1.1 PLAN RETRIEVAL, ADJUSTMENT, AND STORAGE

Our basic plan-retrieval algorithm is quite simple: we scan the plan library, matching forms in the library plan's goal with the input goal. We retrieve the library plans with the greatest number of matches, then break ties by counting the number of matches between the input initial conditions and the initial conditions of the tied plans. Ties in the number of matches for both goal and initial expressions are broken arbitrarily.





This process selects a single library plan, but its initial and goal conditions need not match the input initial and goal expressions exactly. The *adjustment* process adds goals to the library plan that appear in the input goal but are not already in the plan, and deletes goals from the library plan that do not appear in the input goal expression. The library plan's initial conditions are changed similarly to match the input problem description. Then causal links are adjusted: a link in the library plan of the form $S_i \xrightarrow{Q} S_j$ where $S_i$ has been deleted becomes an open condition of the form $\xrightarrow{Q} S_j$; if $S_j$ has been deleted, the link itself can be removed. New open conditions are also added for any new goal forms. This new plan is guaranteed to be a refinement of the null plan for the current problem, but unlike the library plan it is not necessarily complete. See Section 6 for more details on the retrieval and adjustment algorithms.

Then the adaptation phase is initiated, which modifies the retrieved plan to produce a solution plan for the new problem. This solution is passed to the library *storage* routine, which decides whether and how to store the plan for use in subsequent planning episodes. The question of *whether* to store a newly adapted solution back into the plan library is an important one, since having more plans in the plan library makes the library-retrieval process take longer. On the other hand, storing many plans in the library increases the chances that one will be a close match to a subsequent input problem.

Ideally the plan library should consist of a relatively small set of "qualitatively different" solutions to "commonly occurring" problems, but a characterization of qualitatively different and of commonly occurring can be hard to come by. SPA makes no contribution to the question of what should appear in the plan library, and our empirical work in Section 8 assumes a predetermined plan library which is not augmented during the experimental trials. See (Veloso, 1992) for an illuminating investigation of these issues.

## 4.2 The Adaptation Loop

The **AdaptationLoop** function is similar to its generative counterpart **RefinementLoop** except in the latter case every plan selected for refinement is further refined. In the case of adaptation, a partial plan might be marked for refinement or alternatively for *retraction*, and the algorithm must keep track of which. Thus the frontier becomes a set of pairs of the form $\prec \mathcal{P}, d \succ$ where $\mathcal{P}$ is a partial plan and $d$ is a symbol denoting a *direction*, either DOWN or UP. The DOWN case means refine the plan further, in which case the **RefinePlan** function is called, exactly the same as in generation. A direction of UP results in a call to **RetractPlan**, which is defined below.

**function AdaptationLoop**(InitialPlan): Plan or *failure*

| | |
|---|---|
| 1 | Frontier :=$\{\prec$InitialPlan, UP$\succ$, $\prec$InitialPlan, DOWN$\succ\}$ |
| 2 | **loop forever:** |
| 3 |     **if** Frontier is empty **then return** *failure* |
| 4 |     $\prec$P, D$\succ$ := select an element from Frontier |
| 5 |     Delete $\prec$P, D$\succ$ from Frontier |
| 6 |     **if** P is a solution **then return** P |
| 7 |     **if** D = DOWN **then** |
| 8 |         **for** each plan $P_i$ returned by RefinePlan(P) **do** |





9                  Add $\prec P_i$, DOWN$\succ$ to `Frontier`
10       **else if** D = UP **then**
11            add all elements of `RetractRefinement(P)` to `Frontier`

## 4.3 Retracting Refinements

Instead of adding and protecting causal links, retraction removes choices made when the library plan was originally generated. Just as `RefinePlan` selects a flaw in the current plan and adds to the frontier all different ways of fixing the flaw, `RetractRefinement` takes a prior refinement choice, uses the associated reason structure to completely remove that refinement, and adds to the frontier all of the alternative ways that the refinement might have been made.

As Figure 3 illustrates, retraction replaces a queue entry of the form $\prec \mathcal{P}$, UP$\succ$ with a "parent" of $\mathcal{P}$'s (also tagged UP) along with a set of $\mathcal{P}$'s siblings, each tagged DOWN. A precise definition of "sibling" is the set of refinements to $\mathcal{P}$'s parent that are not *isomorphic to* $\mathcal{P}$. We define isomorphism as follows:

**Definition:** *Two plans $\mathcal{P}_1$ and $\mathcal{P}_2$ are isomorphic just in case*

1. *Steps agree:*

   - *there is a 1:1 mapping from steps in $\mathcal{P}_1$ and $\mathcal{P}_2$ such that corresponding steps have identical names (take the correspondence to be $S_1, S_2, \ldots S_n$ to $R_1, R_2, \ldots R_n$)*

2. *Links agree:*

   - $S_i \xrightarrow{Q} S_j \in \mathcal{P}_1$ *iff* $R_i \xrightarrow{Q} R_j \in \mathcal{P}_2$

3. *Orderings agree:*

   - $S_i < S_j \in \mathcal{P}_1$ *iff* $R_i < R_j \in \mathcal{P}_2$

4. *Binding constraints agree:*

   - $(= ?\mathtt{s}_i\ \mathtt{K}) \in \mathcal{P}_1$ *iff* $(= ?\mathtt{r}_i\ \mathtt{K}) \in \mathcal{P}_2$, *where* $?\mathtt{s}_i$ *is a variable in step $i$ of $\mathcal{P}_1$ and* $?\mathtt{r}_i$ *is the corresponding variable in step $i$ of $\mathcal{P}_2$ and K is a constant*
   - *likewise for* $(\neq ?\mathtt{s}_i\ \mathtt{K})$ *and* $(\neq ?\mathtt{r}_i\ \mathtt{K})$
   - $(= ?\mathtt{su}_i\ ?\mathtt{sv}_j) \in \mathcal{P}_1$ *iff* $(= ?\mathtt{ru}_i\ ?\mathtt{rv}_j) \in \mathcal{P}_2$, *where* $?\mathtt{su}_i$ *and* $?\mathtt{sv}_j$ *are variables in steps $i$ and $j$ of $\mathcal{P}_1$ respectively, and* $?\mathtt{ru}_i\ ?\mathtt{rv}_j$ *are the corresponding variables in steps $i$ and $j$ of $\mathcal{P}_2$ respectively*
   - *likewise for* $(\neq ?\mathtt{su}_i\ ?\mathtt{sv}_j)$ *and* $(\neq ?\mathtt{ru}_i\ \mathtt{rv}_j)$.

This definition implies that two isomorphic plans have the same open conditions and threatened links as well. Note that two plans may have corresponding steps and identical orderings and *not* be isomorphic, however, since they can differ on one or more causal links.





The question now arises as to which decisions can be reversed when moving upward in the space of partial plans. The simplest answer is that `RetractRefinement` must be able to eliminate any decision that could have been made by `RefinePlan`. Refinement decisions made by `RefinePlan` can result in the following elements being added to a plan:

- A single causal link, plus an ordering constraint plus binding constraints inserted to fix an open condition. In this case all the constraints will be tagged with the reason [ESTABLISH $S_i \xrightarrow{Q} S_j$].

- A new step plus a causal link, inserted to fix an open condition. In this case two ordering constraints and a set of binding constraints associated with the step will be tagged with the reason [ADD-STEP $S$], and an ordering constraint and a second set of binding constraints will be added along with the new link, as above.

- An ordering constraint inserted to fix a threat either by promotion or demotion. This constraint will be tagged with [PROTECT $\prec S_i \xrightarrow{Q} S_j, S_t \succ$] where $S_t$ is the threatening step.

- A set of variable-binding constraints plus two ordering constraints inserted to fix a threat by separation. These constraints will be tagged with [PROTECT $\prec S_i \xrightarrow{Q} S_j, S_t \succ$].

A single call to `RetractRefinement` should therefore retract one such refinement decision, which amounts to removing the associated set of orderings, binding constraints, steps, and links from the plan. Notice that a decision corresponds closely to a set of identical *reason* structures in a plan, so retracting a decision from a plan really amounts to removing a set of constraints with identical tags, along with their associated links and steps.

The one exception to this correspondence is the fact that the decision to add a step to a plan (reason [ADD-STEP ...]) is *always* made as part of a decision to add a link (reason [ESTABLISH ...]), so these two decisions should be retracted as a pair as well. We will treat the two as separate decisions, but our algorithm will ensure that a step is removed from a plan as soon as its last causal link is retracted.

Although the choice of a decision to retract is made nondeterministically, it cannot be made arbitrarily, since the planner could not have generated the decisions in any order. For example, when building plan $\mathcal{P}$, the planner might have created a link $S_i \xrightarrow{Q} S_j$ and later introduced a set of ordering or binding constraints $C$ to protect this link from being threatened by another step $S_t$. The retraction algorithm must be able to retract either decision (delete the link or the constraints), but these two decisions are not symmetric. If $C$ is deleted, $L$ becomes threatened again, but if $L$ is deleted, then $C$ becomes superfluous.

To protect against leaving the plan with superfluous steps, links, or constraints, we allow the algorithm to retract only those decisions that are *exposed*. Informally, a decision is exposed if no other constraints in the plan depend on the structure added to the plan by that decision. The formal definition of exposed is stated in terms of *reasons* within a plan, since as we noted above decisions add constraints to a plan that are tagged with identical reasons.





**Definition:** *A reason* $R$ *is* exposed *in plan* $\mathcal{P}$ *if*

1. $R$ *is of the form* $[\text{PROTECT } S_i \xrightarrow{Q} S_j]$ *for some link* $S_i \xrightarrow{Q} S_j$, *or*

2. $R$ *is of the form* $[\text{ESTABLISH } S_i \xrightarrow{Q} S_j]$ *for some link* $S_i \xrightarrow{Q} S_j$ *and*

   (a) $\mathcal{P}$ *contains no reason of the form* $[\text{PROTECT } S_i \xrightarrow{Q} S_j]$, *and*

   (b) *either* $S_i$ *participates in another link* $S_i \xrightarrow{Q} S_x$, *or* $S_i$ *does* not *appear in any protected threat of the form* $[\text{PROTECT} \prec S_x \xrightarrow{Q} S_y, S_i \succ]$, *or*

3. $R$ *is of the form* $[\text{ADD-STEP } S_k]$ *and* $\mathcal{P}$ *contains no link of the form* $S_k \xrightarrow{Q} S$.

The first and third cases are fairly straightforward: constraints that resolve a threat can always be retracted, and a step can only be removed if it no longer participates in any causal links.

The second case deserves some explanation, however. The first subcase says that a link cannot be deleted from a plan as long as there are constraints in the plan protecting it from a threat—otherwise the constraints added to resolve the threat would become superfluous. The second subcase guards against the following special case: suppose that $\mathcal{P}$ contains only two links, $S_i \xrightarrow{Q} S_j$ and $S_x \xrightarrow{R} S_y$. Furthermore, suppose that $S_i$ posed a threat to $S_x \xrightarrow{R} S_y$, but a previous decision resolved that threat. One might be tempted to remove the link $S_i \xrightarrow{Q} S_j$ and along with it the step $S_i$, since $S_i$ would no longer serve any purpose in the plan. But doing so would leave superfluous structure in the plan, namely the constraints that were added to resolve the threat $\prec S_x \xrightarrow{R} S_y, S_i \succ$. Our definition for exposed guarantees first that a step will be removed whenever it ceases to serve a purpose in the plan's causal structure (i.e. whenever its last link is removed), but that doing so will never leave superfluous constraints in the plan.

Now the order in which decisions can be retracted can be stated simply: a decision can be retracted only if its associated reason is exposed. Obeying this ordering means that the plan will never contain superfluous constraints, links, or steps; equivalently we might say that retracting only exposed decisions corresponds to the reverse order in which a generative planner *might have made* those decisions originally.

Constraining retraction to occur in this order might seem to be overly restrictive, so we make two important observations. First, note that the order of retraction is *not* constrained to be the reverse of the order used when the library plan was created — only the reverse of *one* of the decision-orderings that *could* have been used to create the library plan. Second, we direct the reader to Section 7, which explains how CHEF repair strategies, encoded as SPA heuristics, could sidestep these restrictions by acting as macro operators.

Next we present the `RetractRefinement` function. Notice how the definition mirrors that of its generative counterpart `RefinePlan`: the latter chooses a flaw and returns a list that includes all possible ways of fixing it, the former chooses an exposed decision, *removes* the constraints that originally fixed it, and enqueues all the alternative ways of fixing it.





**function RetractRefinement(P)**: List of ≺Plan, Direction≻

1    R := select an exposed reason
2    **if** there is no exposed reason **then return** {}.
3    (F, P′) := RemoveStructure(R, P)
4    **collect**≺P′, UP≻
5    **for** each plan P″ returned by CorrectFlaw(F,P′) **do**
6        **if** P″ is not isomorphic to P **then collect**≺P″, DOWN≻
7    **return** all plan, direction pairs collected in lines 4 and 6.

The way to remove the structure associated with an exposed reason depends on the type of the reason. The function RemoveStructure returns the flaw associated with the input reason as well as the plan produced by removing the appropriate constraints, links, and steps. Notice that the coupling between link and step decisions is made here: when the last link to a step is deleted the step is deleted too. For this reason we do not have to handle the case of removing a reason of the form [ADD-STEP S]: a step becomes exposed only when a link is deleted, but this function removes the step immediately. So a reason of the form [ADD-STEP S] will never appear exposed in a plan.

**function RemoveStructure(R, P)**: (Flaw, Plan)

1    **if** R is of the form [PROTECT ≺$S_i \xrightarrow{Q} S_j, S_t$≻] **then**
2        F := ≺$S_i \xrightarrow{Q} S_j, S_t$≻
3        P′ := a copy of P
4        Delete from P′ all constraints tagged with R
5        **return** (F, P′)
6    **else if** R is of the form [ESTABLISH $S_i \xrightarrow{Q} S_j$] **then**
7        F := $\xrightarrow{Q} S_j$
8        P′ := a copy of P
9        Delete $S_i \xrightarrow{Q} S_j$ from P′
10       Delete from P′ all constraints tagged with R
11       **if** P′ contains no link of the form $S_i \xrightarrow{Q} S_k$ for any step $S_k$ and expression $Q$ **then**
12           delete $S_i$ from P″ along with all constraints tagged with [ADD-STEP $S_i$]
13       **return** (F, P′)

This concludes the description of the SPA algorithm; we next examine the algorithm's formal properties, proving that it is sound (any plan it returns constitutes a solution to the input planning problem), complete (if there is *any* solution to the input planning problem, SPA will eventually find it, regardless of the library plan it chooses to adapt), systematic (the adaptation will never consider a partial plan more than once).

## 5. Soundness, Completeness, and Systematicity

To prove formal properties of the SPA algorithm we begin by characterizing a lifted version of the generative algorithm developed by McAllester and Rosenblitt's (1991) algorithm





(hereafter called SNLP) in terms of a search through the space of partial plans. We then consider retraction as well. This discussion uses many of the concepts and terms from Section 3 describing plans and planning problems.

Consider a directed graph as in Figure 1 where a node represents a plan and an arc represents a plan-refinement operator. We can define the children of a node (plan) $\mathcal{P}$, subject to a nondeterministic choice, as follows:

**Definition:** *The children of a plan $\mathcal{P}$ are exactly these:*

1. *If $\mathcal{P}$ is complete then it has no children.*

2. *Otherwise select one of $\mathcal{P}$'s open conditions or threatened links.*

3. *If the choice is the open condition, $\xrightarrow{Q} S_j$, then $\mathcal{P}$'s children are all plans that can be constructed by adding a link $S_i \xrightarrow{Q} S_j$, an ordering $S_i < S_j$, and a minimal variable binding constraint $\theta$, where $S_i$ is either an existing step or a newly created step that can consistently be ordered prior to $S_j$, and that adds some proposition R, where $R\theta = Q$.*

4. *Otherwise, if the choice is the threat, $\prec S_i \xrightarrow{Q} S_j,\ S_t \succ$, then the node has the children obtained by*

   (a) *adding the ordering $S_t < S_i$*

   (b) *adding the ordering $S_j < S_t$*

   (c) *adding the orderings $S_i < S_t$ and $S_t < S_j$ in addition to a minimal variable binding constraint, $\theta$, that forces all forms R in $S_t$'s add and delete list, $R\theta$ doesn't unify with Q.*

   *provided these are consistent with the constraints currently in $\mathcal{P}$.*

McAllester and Rosenblitt (1991) claim three properties of this representation and algorithm:

- Soundness: a leaf node corresponds to a partial plan, any completion of which will in fact satisfy the input goal.

- Completeness: *any* plan that solves the planning problem is realized in the graph as a leaf node. Therefore any strategy for searching the graph that is guaranteed to consider every node eventually will find a solution to the planning problem if one exists.

- Systematicity: two distinct nodes in the graph represent non-isomorphic plans, and furthermore, the graph generated by a planning problem is a tree. Therefore a search of the plan graph that does not repeat a node will never consider a partial plan or any of its refinements more than once.





### 5.1 Soundness

The soundness property for SPA follows directly from SNLP's soundness, since soundness is not a property of the algorithm's search strategy, but comments only on the nature of leaf nodes (complete plans). Since SPA defines plans and solutions in the same way as SNLP, SPA too is sound.

### 5.2 Completeness

Completeness, recall, consists of two claims:

1. that every solution to the planning problem is realized as a leaf node of the graph, and

2. that the search algorithm will eventually visit every leaf node in the graph.

The first condition once again does not depend on the way the graph is searched, therefore it is true of SPA because it is true of SNLP. The second condition is less clear, however: SNLP makes sure it covers the entire graph by starting at the root and expanding the graph downward in a systematic fashion, whereas SPA starts at an arbitrary point in the graph and traverses it in both directions.

A proof of completeness amounts to demonstrating that for any partial plan $\mathcal{P}_i$ representing the beginning point for SPA—the case (library plan) supplied by the retrieval mechanism—the algorithm will eventually retract constraints from the plan until it visits the root node (null plan), and doing so also implies that it will visit all subtress of the root node as well. More formally stated, we have:

**Theorem 1:** *A call to* `AdaptPlan` *with a library plan $\mathcal{P}$ will cause every partial plan (every node in the plan graph defined by $\mathcal{P}$'s planning problem) to be visited.*

We use an inductive argument to prove this theorem, showing that the subgraph rooted at $\mathcal{P}_i$ is completely explored, and that the algorithm will follow a path up to the root (null plan) exploring every subgraph in the process.

We begin by demonstrating informally that SPA's method of *refining* a partial plan (adding constraints as opposed to retracting) is equivalent to the graph search undertaken by SNLP. (Recall that SPA operates by manipulating a search frontier whose entries are $\prec\mathcal{P}$, DOWN$\succ$ and $\prec\mathcal{P}$, UP$\succ$, corresponding respectively to adding and deleting constraints from $\mathcal{P}$.)

**Claim 1:** *The entries generated by SPA's processing an entry of the form $\prec\mathcal{P}$, DOWN$\succ$ correspond exactly to the SNLP graph of partial plans rooted at $\mathcal{P}$, assuming the same choice is made as to what condition (open or threat) to resolve at each stage.*

It suffices to show that the new entries generated by SPA in response to an entry of the form $\prec\mathcal{P}$, DOWN$\succ$ correspond to the same partial plans that comprise $\mathcal{P}$'s children in the graph as defined above (Page 337). There were three parts to the definition: $\mathcal{P}$ complete, $\mathcal{P}$ refined by choosing an open condition to satisfy, $\mathcal{P}$ refined by choosing a threat to resolve.

In the case that $\mathcal{P}$ is complete, $\mathcal{P}$ has no children, and likewise SPA terminates generating no new entries.





Otherwise SPA calls `RefinePlan`, which chooses a condition to resolve and generates new DOWN entries, one for each possible resolution. Note therefore that a DOWN entry generates *only* DOWN entries; in other words refinement will only generate more refinements just as a directed path in the graph leads to successively more constrained plans.

In the second case an open condition is chosen; `RefinePlan` generates new DOWN entries for all existing steps possibly prior to the open condition and for all actions that add the open condition's proposition. This corresponds exactly to case (3) above.

In the last case a threat condition (a link and a threatening step) is chosen; `RefinePlan` adds the orderings and/or binding constraints that prevent the threat, exactly as in case (4) above.

Having verified that SPA generates the immediate children of a partial plan in a manner equivalent to SNLP, and furthermore having noted that it enters these children on the frontier with DOWN tags as well (so their children will also be extended), the following lemma follows directly from Claim 1 above, the completeness of SNLP, and a restriction on the search algorithm noted below:

**Lemma 1:** *If* SPA *ever adds to the frontier the entry* $\prec \mathcal{P}$, DOWN$\succ$ *then it will eventually explore all partial plans contained in the graph rooted at* $\mathcal{P}$ *(including* $\mathcal{P}$ *itself).*

One must be precise about what it means to "explore" a partial plan, or equivalently to "visit" the corresponding graph node. `AdaptPlan` contains a loop in which it selects an entry from the frontier (i.e. a plan / direction pair), checks it for completeness (terminating if so), and otherwise refines the plan. So "exploring" or "considering" a plan means selecting the plan's entry on the search frontier. Lemma 1 actually relies on a search-control strategy that is guaranteed eventually to consider every entry on the frontier. This corresponds to a search strategy that will eventually visit every node in a graph given enough time—in other words, one that will not spend an infinite amount of time in a subgraph without exploring other areas of the graph. SNLP's iterative-deepening search strategy has this property as does SPA's breadth-first search.

The base case for completeness follows directly from Lemma 1 and the fact that `AdaptPlan` initially puts both $\prec \mathcal{P}_i$, UP$\succ$ and $\prec \mathcal{P}_i$, DOWN$\succ$ on the frontier:

**Lemma 2:** *The subgraph rooted at* $\mathcal{P}_i$ *will be fully explored.*

Now we can state the induction condition as a lemma:

**Lemma 3:** *If a partial plan* $\mathcal{P}$ *is fully explored, and* $\mathcal{P}_p$ *is the partial plan generated as a result of (nondeterministically) retracting a choice from* $\mathcal{P}$, *then the subgraph rooted at* $\mathcal{P}_p$ *will be fully explored as well.*

The fact that $\mathcal{P}_p$ is considered as a result of a retraction from $\mathcal{P}$ means that the entry $\prec \mathcal{P}$, UP$\succ$ was considered, resulting in a call to `RetractRefinement` from which $\mathcal{P}_p$ was generated as the parent node $\mathcal{P}'$ in the call to `RetractRefinement`. To show that $\mathcal{P}_p$'s subgraph is fully explored we need to show that

1. $\mathcal{P}_p$ is visited,

2. the subgraph beginning at $\mathcal{P}$ is fully explored, and

3. that all of $\mathcal{P}_p$'s children *other* than $\mathcal{P}$ are fully explored.





The first is true because `RetractRefinement` generates the entry $\prec \mathcal{P}_p$, UP$\succ$, which means that $\mathcal{P}_p$ will eventually be visited. The second condition is the induction hypothesis. The third condition amounts to demonstrating that (1) the children returned by `Retract-Refinement` actually represent $\mathcal{P}_p$'s children as defined above, and (2) that these children will themselves be fully explored.

The first is easily verified: `RetractRefinement` immediately calls `CorrectFlaw` on the flaw it chooses to retract, which is exactly the function called by `RefinePlan` to address the flaw in the first place. In other words, the new nodes generated for $\mathcal{P}_p$ by `RetractRefinement` are exactly those that would be generated by `RefinePlan`, which by Claim 1 are $\mathcal{P}_p$'s children.

As for the children being fully explored, all the children except for $\mathcal{P}$ itself are put on the frontier with a DOWN tag, and therefore by Lemma 1 will be fully explored. $\mathcal{P}$ itself is fully explored by assumption, which concludes the proof of Lemma 3.

Finally we need to demonstrate the call to `AdaptPlan`($\mathcal{P}_i$) eventually retracts to the graph's root. First of all, the first call to `AdaptPlan` generates an entry of the form $\prec \mathcal{P}_i$, UP$\succ$, and processing an entry of the form $\prec \mathcal{P}_i$, UP$\succ$ generates an entry of the form $\prec \mathcal{P}_{i+1}$, UP$\succ$, where $\mathcal{P}_{i+1}$ represents the retraction of a single constraint from $\mathcal{P}_i$.

The call to `AdaptPlan`($\mathcal{P}_i$) therefore generates a sequence of entries of the form $\prec \mathcal{P}_1$, UP$\succ$, $\prec \mathcal{P}_2$, UP$\succ$, ..., $\prec \mathcal{P}_k$, UP$\succ$, where $k$ is the number of decisions[6] in $\mathcal{P}_i$. In this sequence $\mathcal{P}_1 = \mathcal{P}_i$ and $\mathcal{P}_k$ has no constraints. Furthermore, Lemma 2 tells us that the subgraph rooted at $\mathcal{P}_1$ is fully explored and Lemma 3 tells us that the rest of the $\mathcal{P}_i$ subgraphs are fully explored as well.

The final question is whether $\mathcal{P}_k$, a plan with no constraints, is necessarily the null plan (defined above to be a plan with just the initial and final steps and the single constraint ordering initial before final). We know that calls to `RetractRefinement` will eventually delete all causal links and all orderings that were added as the result of protecting a threat. Superfluous steps (steps that have no associated link) and orderings (that were added without a corresponding threat condition) might appear in $\mathcal{P}_i$, however, and `RetractRefinement` would never find them. $\mathcal{P}_k$, then, would contain no more retraction options, but would not be the null plan.

We can fix this easily enough, either by requiring the library-retrieval machinery to supply plans without superfluous steps and constraints, or by inserting an explicit check in `RetractRefinement` that removes superfluous steps and constraints when there are no more options to retract.

The former might not be desirable: the library plan might contain steps that don't initially appear to serve the goal, but later come in handy; leaving them in the plan means the planner need not re-introduce them into the plan. The latter option is inexpensive, and is actually implemented in our code. See Section 6.3 for further discussion of this issue.

Assuming that $\mathcal{P}_k$ is the null plan, the completeness proof is finished: we showed that calling `AdaptPlan`($\mathcal{P}_i$) fully explores its own subgraph, and furthermore generates a path to the graph's root (the null plan) ensuring that all nodes below the path are visited in the process.

---

6. More precisely, the number of distinct *reason* structures.





## 5.3 Systematicity

Systematicity, like completeness, is a two-part claim. The first is formal: that the plan graph is a tree—in other words, that the policy of generating a node's parents by making a nondeterministic but fixed choice of a condition (open or threat) to resolve, then generating the node's children by applying all possible ways to resolve that condition means that any two distinct plan nodes represent non-isomorphic plans. The second claim is that the strategy for searching the graph never visits a plan node more than once.

The first claim applies just to the formal definition of the plan graph, so the systematicity of SNLP suffices to prove the systematicity of SPA.

To verify the second claim we need only to show that for any partial plan $\mathcal{P}$, SPA will generate that plan just once. We demonstrate this in two parts:

**Lemma 4:** *Processing an entry of the form $\prec\mathcal{P}$, DOWN$\succ$ will never cause $\mathcal{P}$ to be generated again.*

This is true because generating $\prec\mathcal{P}$, DOWN$\succ$ causes $\mathcal{P}$'s children to be generated with DOWN tags, and so on. Every successive node that gets generated will have strictly more constraints or more links than $\mathcal{P}$, and therefore will not be isomorphic.

**Lemma 5:** *Processing an entry of the form $\prec\mathcal{P}$, UP$\succ$ will never cause $\mathcal{P}$ to be generated again.*

Processing $\prec\mathcal{P}$, UP$\succ$ causes $\mathcal{P}$'s parent $\mathcal{P}_p$ to be generated with an UP tag and $\mathcal{P}$'s siblings to be generated with a DOWN tag. Note that $\mathcal{P}$ is *not* generated again at this point.

No further extension of a sibling of $\mathcal{P}$ can ever be isomorphic to $\mathcal{P}$, since they will differ (at least) on the selection of a solution to the condition resolved between $\mathcal{P}_p$ and its children. Likewise, no sibling of $\mathcal{P}_p$ can ever be refined to be isomorphic to $\mathcal{P}$, since it will differ from $\mathcal{P}$ (at least) in the constraint that separates $\mathcal{P}_p$ from its siblings.

Therefore as long as a plan is not explicitly entered on the frontier with both DOWN and UP tags, it will never be considered more than once. Actually the fitted library plan, $\mathcal{P}_i$, is initially entered on the queue with both DOWN and UP tags, so SPA may consider this partial plan more than once, and is therefore not strictly systematic. Every other partial plan, however, is generated during an iteration of the loop in `AdaptPlan`, which generates each of its plans only once, either UP or DOWN. So the SPA graph-search algorithm is systematic except for the fact that it might consider its initial plan twice.

## 6. Interactions between Retrieval and Adaptation

While the bulk of our research has been devoted to the adaptation phase of the planning process, it is impossible to consider this phase completely in isolation. In this section we consider the expected benefit of adaptation as well as some subtle interactions between adaptation and retrieval. First we compare the complexity of plan adaptation with that of plan generation from scratch; this ratio provides an estimate of how close the library plan must match the current situation in order for adaptation to be faster than generation. Next we outline how plans are stored in and retrieved from SPA's library. Finally we describe some interesting interactions between the processes of retrieval and adaptation.





## 6.1 Should one adapt?

All planners that reuse old cases face the fundamental problem of determining which plans are "good" to retrieve, i.e., which plans can be cheaply adapted to the current problem. In this section we present a simple analysis of the conditions under which adaptation of an existing case is likely to be more expeditious than generative planning.

The basic idea is that at any node, adaptation has exactly the same options as generative planning, plus the opportunity to retract a previous decision. Thus the search space branching factor is one greater for adaptation than for generative planning.

Suppose that the generative branching factor is $b$ and a working plan of length $n$ exists for the new problem. In this case, the cost of generation is $b^n$. Now suppose that the library-retrieval module returns a plan that can be extended to a working plan with $k$ adaptations; this corresponds roughly to the addition of $k$ new steps or the replacement of $\frac{k}{2}$ inappropriate steps. Thus adaptation is likely to be faster than generative planning whenever

$$(b + 1)^k < b^n$$

This inequality is satisfied whenever

$$\frac{k}{n} < \log_{b+1} b$$

As the branching factor $b$ increases, the logarithm increases towards a limit of one. Thus small branching factors exact the greatest bound on the $\frac{k}{n}$ ratio. But since generative planning almost always has a branching factor of at least 3 and since $\log_4 3 = 0.79$, we conclude that adaptation is likely preferable whenever the retrieval module returns a case that requires at most 80% as many modifications as generative planning would require. A conservative estimate suggests that this corresponds to a fitted library plan in which at most 40% of the actions are inappropriate. While we acknowledge that this analysis must be taken loosely, we believe it provides useful intuitions on the case-quality required to make adaptation worthwhile.

## 6.2 The retrieval phase

Our model of retrieval and adaptation is based on the premise that the SPA algorithm itself generates its library plans. Plans generated by SPA automatically have stored with them all the dependencies introduced in the process of building the plan, i.e. all of its causal links and constraints.

Most of a plan's propositions are variabilized before the plan is stored in the library—we do the variabilization in a problem-specific manner, but the general issue of what parts of a plan to variabilize can be viewed as a problem of explanation-based generalization, and is discussed by Kedar-Cabelli and McCarthy (1987) and by Kambhampati and Kedar (1991).

Library retrieval is a two-step process: given a set of initial and goal conditions, the algorithm first identifies the most promising library plan, then does some shallow modification to make the plan's initial and goal conditions match the inputs.

### 6.2.1 Library retrieval

The first phase of the retrieval process uses either an application-supplied method or a domain-independent algorithm similar to the one used by Kambhampati and Hendler (1992) to select candidate plans. First the input goals are matched against the each library plan's goals, and the libary plans with the greatest number of matches are identified. This can result in many candidates, since several plans can match, and a single plan can match in a number of different ways. To choose among the remaining alternatives the algorithm examines the match between the initial conditions. It computes for each alternative the number of open conditions created by replacing the library plan's initial conditions with the input initial conditions. This is intended to measure the amount of planning work necessary to get the input initial world state to the state expected by the library plan. It counts the number of open conditions for each option and chooses the plan with the minimum, breaking ties arbitrarily.

### 6.2.2 Fitting the retrieved plan

Having matched a library plan, fitting it to the new problem is simple:

1. Instantiate the library plan with the variable bindings produced by the match above.

2. Replace the library plan's goal conditions with the new goal conditions.

3. Create a new open condition for each goal proposition that appears in the new goal set but not in the library plan's goal set.

4. Replace the library plan's initial conditions with the new problem's initial conditions.

5. For each causal link that "consumes" a proposition from the old initial conditions, if that proposition is absent from the new initial conditions, then delete the link and add a corresponding new open condition.

6. For each causal link that "produces" a proposition for the old goal conditions, if that proposition is absent from the new goals, then delete the link.

## 6.3 Conservative vs. generous fitting

The algorithm above does no pruning of superfluous steps: the plan returned can contain steps that existed to "produce" causal links for propositions in the library plan's goal set that are not part of the new goals. Hence the fitted plan can contain links, steps, and constraints which are (apparently) irrelevant to the current problem. Of course, until the adaptation algorithm actually runs it is impossible to tell whether these parts of the library plan will actually turn out to be useful. If removed during the fitting process, the adaptation algorithm might discover that it needs to re-generate the same structures.

The question therefore arises as to whether the fitting algorithm should delete all such links, potentially removing many steps and constraints (a conservative strategy), or should it leave them in the plan hoping that they will eventually prove useful (a generous approach)? One can easily construct cases in which either strategy performs well and the other performs poorly.

We noted above an interesting interaction between the generous strategy and our adaptation algorithm. `AdaptPlan`'s retraction algorithm is the inverse of extension, which means that it can only retract decisions that it might have actually made during extension. `AdaptPlan` will obviously never generate a superfluous plan step, and so a library plan containing superfluous links or steps could not have been produced directly by the adapter. If so, `AdaptPlan` might not be able to retract all the planning decisions in the library plan, and is therefore not complete. (Since it cannot retract all previous planning decisions, it cannot retract all the way back to the null plan, and therefore may fail to explore the entire plan space.) Recall from Section 5.2 that the retraction algorithm presented in Section 4 is only complete when used in conjunction with a conservative fitting strategy, or alternatively by modifying the `RetractRefinement` code so it deletes superfluous steps—steps other than the initial and goal steps that do not produce a causal link—from any plan it returns.

## 7. Transformational Adaptation

Most previous work on case-based planning has concentrated on finding good indexing schemes for the plan library, with the idea that storing and retrieving appropriate cases would minimize the need for adaptation. We can nonetheless use the SPA framework to analyze the adaptation component of other systems. The repair strategies included in the CHEF system (Hammond, 1990), for example, specify transformations that can be decomposed into sequences of SPA *refine* and *retract* primitives. Our analysis proves useful in two different ways:

1. It shows how CHEF's indexing and repair strategies could be exploited in the SPA framework by providing heuristic search-control information.

2. It demonstrates how SPA's simple structure can be used to analyze more complex adaptation strategies, and ultimately could be used to compare alternative theories of plan repair.

We start with a section summarizing CHEF's design. Then in Section 7.2 we consider its repair strategies sequentially, decomposing them into SPA operators. Section 7.3 proves that CHEF's set of repairs is incomplete, and Section 7.4 discusses ways to encode CHEF's heuristics in SPA's framework. Section 7.5 discusses how our analysis could be extended to other transformational planners such as GORDIUS (Simmons, 1988).

### 7.1 Plan adaptation in CHEF

CHEF uses a five-stage process for adapting an existing plan to achieve new goals. CHEF first takes a library plan, fits it to the new problem, and simulates its execution, using the new initial conditions and goals. Roughly speaking, CHEF's failures correspond to a SPA plan with at least one flaw—a threatened link or open precondition.

CHEF next uses forward and backward chaining to analyze the failure, discovering things like what step or steps caused the failure, and what goals those steps were servicing. The result is a causal network corresponding to the causal links constructed by SPA in the process of plan generation.[7] SPA therefore performs the first two stages of CHEF's adaptation process in the process of plan generation.

CHEF then uses the causal explanation to select one of sixteen prestored diagnoses, called TOPs (Schank, 1982). A TOP also contains a set of repair strategies—plan transformations that might eliminate the failure—and each repair strategy has an associated test to check its applicability. CHEF's TOPS are divided into five classes: failures due to side effects, desired effects, side features, desired features, and step parameters. The STRIPS action representation used by SPA does not distinguish between object features and other propositions and does not allow parameterized steps, so only the first two classes of TOPs are relevant to our analysis. In any case, these two classes are the most important, since they account for more than half of the TOPs. The distinction between a side effect and a desired effect is straightforward: side effects are operator postconditions that don't support a link, while desired effects do have a purpose in the plan. Naturally, the set of appropriate repairs are different in the two cases.

After choosing a TOP, CHEF instantiates the associated repair strategies using the details of the current planing task. For each possible repair CHEF runs a test to see if the repair is applicable, using the result of this test to instantiate the repair. For example, the test for an abstract repair corresponding to insertion of a "white knight" (Chapman, 1987) would determine which steps could reassert the desired proposition.

Finally, CHEF uses a set of heuristic rules to rank the various instantiated repairs, chooses the best, and applies it. Once the plan is transformed, it is simulated once again; detection of a new failure starts the cycle anew.

## 7.2 Plan transformation in CHEF

Seven of CHEF's seventeen repair strategies do not apply to our STRIPS-like action representation. For example, the repair that adjusts the duration of an action is inapplicable since all STRIPSactions are assumed to occur instantaneously. The rest of this section describes the ten relevant repairs and reduces them to SPA primitives.

- Four repairs add new steps to the flawed plan. In each case the plan failure corresponds to a link $S_p \xrightarrow{Q} S_c$ threatened by another step $S_t$.

  1. **Recover**—Add a new step after $S_t$ that will remove the side-effect proposition $\neg Q$ before the consuming step, $S_c$, is executed.

  2. **Alter-feature**—Add a new step that changes an undesired trait into one that matches the goal.

  3. **Remove-feature**— Add a new step that deletes an object's undesired characteristic.

---

7. Some of CHEF's failure explanations are more expressive than SPA's causal links since the latter cannot express metagoals such as avoiding wasteful actions. Here and in the rest of our analysis, we consider only the features of CHEF that pertain to STRIPS planning. We caution the reader that many of CHEF's innovations are relevant only in non-STRIPS domains and thus that our analysis, by necessity, is incomplete.

These three repairs are identical from SPA's perspective, since STRIPS does not distinguish between object features and other types of propositions. Since the repair strategy is the same in the three cases, it is unclear that the distinction CHEF makes between objects and features provides any useful control knowledge in these cases.

Each of these repairs corresponds to the introduction of a "white knight." Accomplishing these repairs systematically requires retracting the threatened link then adding a new link produced by a new step (rather than simply adding the new step). Thus SPA can simulate these three transformations with a retract-refine sequence, although additional retraction might be needed to eliminate decisions that depended on the original threatened link.

4. **Split-and-reform**—Divide a step into two distinct steps that together achieve the desired results of the original.

In SPA terminology, it is clear that the step to be split, $S_p$, must be producing two causal links, since it is accomplishing two purposes. Thus SPA can effect this repair by retracting the threatened link (which automatically removes some variable binding and ordering constraints), adding a new step $S_{p'}$, and then adding a new link $S_{p'} \xrightarrow{Q} S_c$.

- Two transformations replace an existing step in the plan.

5. **Alter-plan:side-effect**—In this case the failure is a link $S_p \xrightarrow{Q} S_c$ which is threatened by another step $S_t$ whose postcondition $\neg Q$ is not involved in another link. The repair is to replace $S_t$ with another step, $S_{t'}$ that doesn't have $\neg Q$ as a postcondition.

6. **Alter-plan:precondition**—This failure is a step $S_c$ which either has an open precondition $Q$ or whose precondition is supported by a threatened link. The repairing transformation replaces $S_c$ with a new step that does not have $Q$ as a precondition.

These transformations have the best potential for providing SPA with search-control heuristics. Both of these repairs make a replacement (*retract* followed by *refine*) to a link in the middle of a causal network. Recall, however, that SPA only makes changes to the "fringe" of the network: SPA only retracts decisions that have no other decisions depending on them. For example, consider the following elaboration of the **Alter-plan:side-effect** example above. Suppose that the current plan contains two additional decisions: the decision to establish a causal link $S_t \xrightarrow{Q} S_u$ (this is what caused the inclusion of $S_t$ to begin with) and also a decision to protect this link from another threatening step $S_k$. Since the latter choice depends on the very existence of the link, the decision to add $S_t$ to the plan as support for $S_t \xrightarrow{Q} S_u$ cannot be retracted until the decision to protect it has been retracted.

Emulation of the **Alter-plan:side-effect** and **Alter-plan:precondition** transformations would result in $n+1$ SPA retract operations followed by $n+1$ refines, where $n$ is the number of dependent decisions in the causal network. In the current SPA implementation, there is no facility for this type of macro operator, but these CHEF transformations suggest the utility of storing the sequence of decisions retracted in the course of bringing $S_c$ to the fringe, and then replaying these decisions with a process similar to derivational analogy (Carbonell, 1983; Veloso & Carbonell, 1993; Veloso, 1992).

- One repair changes the plan's variable-binding constraints.

    7. **Alter-item**—A new object is substituted to eliminate unwanted features while maintaining desired characteristics.

    This repair can be used to correct a number of SPA failures: threatened links, inconsistent constraints, and the inability to support an open condition (due to unresolvable threats). SPA could effect the repair by retracting the decision that added the constraints (most likely the addition of a causal link) and refining a similar decision that bound a new object.

- Three transformations modify the plan's temporal constraints, reordering existing steps.

    8. **Alter-placement:after**—This repair corresponds to promotion and requires no retraction.

    9. **Alter-placement:before**—This repair corresponds to demotion and also requires no retraction.

    10. **Reorder**—The order in which two steps are to be run is reversed. This can be accomplished by retracting the decision that added the original ordering and asserting the opposite ordering.

This analysis of CHEF aids in our understanding of transformational planners in two ways. First it clarifies CHEF's operation, providing a simple explanation of its repair strategies and showing what sorts of transformations it can and cannot accomplish. Second it lays the groundwork for incorporating CHEF's strategies into SPA's adaptation algorithm in the form of control policies (Section 7.4).

## 7.3 The completeness of CHEF

One result of analyzing CHEF's repair strategies in SPA's framework is a demonstration that CHEF's particular set of repairs is incomplete—that is, there are some combinations of library plans and input problems for which CHEF will be unable to generate a suitable sequence of repairs. Consider, for example, the causal structure shown in Figure 4.

Assume that ordering constraints restrict the plan steps to the figure's left-to-right ordering; in other words suppose that the only consistent linearization of this plan starts with $S_a$ then $S_t$ then $S_b$ and so on. The arrows denote causal links, but only two links have been labeled with the proposition produced (the others are irrelevant). Since $S_t$ deletes $P$ and $S_b$ requires it, it is clear that $S_t$ threatens $S_a \xrightarrow{P} S_b$. Since $S_u$ consumes $\neg P$, both $P$ and $\neg P$ are useful effects and the threat must match one of CHEF's desired effect TOPs. In fact, Figure 4 is a classic example of the **blocked-precondition** TOP which has only two repair strategies: **Recover** and **Alter-plan:precondition**. In particular, CHEF is forbidden from trying the **Alter-plan:side-effect** repair since the threat results from a desired effect (for $S_u$) not a side effect. This means that CHEF will never consider replacing $S_t$ with a step that doesn't delete $P$, even though that may be the only way to achieve a working plan. To see that this transformation is capable of resulting in a working plan, note that the choice of $S_u$ to support the goal may have been incorrect. In other words, it may be possible to replace $S_u$ with another step that does not require $\neg P$, which would make the failure a side-effect failure instead of a desired-effect failure, and would enable $S_t$'s replacement.

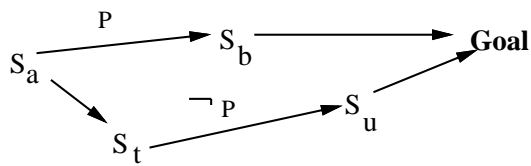

Figure 4: CHEF repairs cannot fix desired-effect plan failures with this causal structure.

What are the implications of this result? Probably CHEF's incompleteness is of minor consequence, especially since that project's goal was to produce a heuristically adequate set of indexing and transformation strategies rather than a formally verifiable algorithm. An analysis like this is nonetheless instructive since it makes precise what tradeoffs CHEF's algorithm makes. It can be instructive to ask *why* a particular algorithm is unsound, incomplete, or unsystematic, and what advantages in expressive power or expected performance are gained by sacrificing some formal property. We believe that an algorithm's formal properties provide one of a number of ways to understand the algorithm's behavior, but do not constitute the ultimate standard by which an algorithm's value should be judged.

We next turn to the topic of how to use the CHEF repair strategies within the SPA framework to guide the adaptation algorithm.

## 7.4 CHEF transformations as SPA heuristics

At the highest level, CHEF and SPA operate in very different ways. CHEF starts with a complete plan that fails to satisfy the goal and uses transformations to generate a new complete plan. CHEF admits no notion of a partial plan and no explicit notion of retracting a commitment. Contrast this with the approach taken by SPA, which can retract any previous planning decision, resulting in an incompletely specified plan. Thus, to endow SPA with search-control heuristics corresponding to CHEF's transformations, we need to chain together SPA's local refine/retract decisions to effect a "jump" from one area of the plan space to another.

The simplest way of giving SPA this capability is to reformulate SPA's top-level control loop from a breadth-first exploration of the space of plans (using a queue or priority queue) to a depth-first or iterative-deepening depth-first search (using a stack). In such a scheme `RefinePlan` would no longer enqueue all the new plans returned by `CorrectFlaw`; instead it would choose the "best" successor plan (using some heuristic ranking information) and explore it, leaving the alternates on the stack for later exploration if backtracking proved necessary. `RetractRefinement` would do likewise with the retracted node's siblings. This modification to SPA's top-level control loop eliminates the need for a global plan-evaluation heuristic, using instead the following four hooks for heuristic control knowledge:

1. When the `RetractRefinement` procedure is given a plan to retract, heuristic information is brought to bear to decide which decision to retract.

2. After `RetractRefinement` generates its new plans it uses heuristics to choose whether to continue retracting constraints from the parent or whether to refine a child (and if it chooses the latter, which sibling to refine).

3. `RefinePlan` likewise uses heuristic information to determine which open condition or threatened link should be addressed first.

4. After `RefinePlan` generates its successor plans it uses heuristics to select which sibling to continue refining.

Consider the operation of this depth-first variant of SPA, given the initial fitted plan tagged both UP and DOWN. Rule 2 applies, since this choice requires deciding between retraction and extension. We could encode CHEF's repair strategies by making Rule 2 examine the current plan's causal structure, use that structure to choose an appropriate TOP, and choose a repair strategy. As described in the previous sections, each repair strategy can be encoded as a macro operator of refines and retracts — these could be written into a globally accessible memory area and "read off" by subsequent rules. A **Recover** repair might expand to a two-step sequence: retract link, refine link. Rule 2 would choose to retract the fitted plan, then Rule 1 would choose the troublesome link to be retracted, then Rule 2 would choose the child corresponding to adding the step specified by the **Recover** repair. At this point, the macro operator would have been completely executed.

Since this new control structure uses only the standard SPA plan modification operators and only returns when the set of open conditions and threatened links are null, soundness is maintained. Similarly, as long as depth-first iterative-deepening search is used, this approach preserves SPA's completeness. Systematicity is violated by the use of iterative-deepening search, however, and there is another problem with systematicity under this approach as well: multiple repairs cannot necessarily be performed in sequence. The latter problem stems from the fact that all CHEF repairs involve refines and most involve retracts followed by refines. Yet, the only plans returned by a call to `RefinePlan` are tagged DOWN and thus they cannot have a transformation involving retraction applied to them (without violating systematicity). There appear to be several possible solutions to this problem:

- Delay attempting any repairs that do not involve retraction, such as **Alter-placement: after** and **Alter-placement:before**, until another repair that *does* retract has been applied.

- Perform all retractions initially, before trying any extension adaptations.[8]

- Ignore the UP and DOWN tags and allow both extension and retraction at any node. While this approach sacrifices systematicity, the hope is that the advantages of search control directed by CHEF-style transformation will offset the increased size of the overall search space. In any case, the approach still guarantees completeness.

---

8. This policy is extremely similar to the adaptation techniques used by PRIAR.

## 7.5 Extending the analysis

This section can only sketch the possibilities for integrating the ideas of transformational planners into the SPA framework. Future research will implement these ideas and test to see whether they work as the previous section suggests. An implementation would also allow ablation studies evaluating the relative utility of different repair heuristics. We suspect that **Alter-plan:side-effect** and **Alter-plan:precondition** would provide the greatest guidance, but we will test this belief.

It would also be interesting to duplicate our analysis of CHEF for other transformational planners. We believe this would be a straightforward exercise in many cases. For example, the first step that GORDIUS (Simmons, 1988) takes when debugging a plan is to build a causal structure like the one SPA builds. Since GORDIUS (like CHEF) uses a rich set of repair heuristics that match faulty causal structures, we suspect that they can be decomposed into SPA-like primitives as were CHEF's. One difficulty in this analysis would concern GORDIUS's emphasis on actions with metric effects. Since SPA's STRIPS representation does not allow conditional effects (nor those computed by arithmetic functions) a first step would be to add SPA-style retraction to the UCPOP (Penberthy & Weld, 1992) or ZENO (Penberthy & Weld, 1993) planners. While UCPOP handles conditional effects and universal quantification, it does not match GORDIUS in expressiveness. ZENO, however, handles metric effects and continuous change.

## 8. Empirical Study

We had two goals in conducting empirical studies:

1. to make more precise the nature and extent of speedup that could be realized by using library-refit planning, and

2. to compare SPA to PRIAR (Kambhampati & Hendler, 1992), which pursues similar ideas within a different planning framework.

The work on PRIAR closely parallels our own: the key idea in both cases is that a generative planning algorithm can be used to adapt library plans, provided that (1) the planner keeps some record of the reasons for its choices, and (2) the planner can retract as well as make refinement choices. Since PRIAR and SPA share use of a STRIPS-like representation, we were able to replicate the experiments undertaken by Kambhampati and Hendler (1992) and compare our results with theirs[9].

## 8.1 Problem statement

First some background: the PRIAR experiments use two general classes of block-stacking problems, named $x$BS and $x$BS1. $x$ is an integer (ranging from 3 to 12) designating the number of blocks involved in that problem.

The first class, e.g. 3BS, involves an initial configuration in which all the blocks are on the table and clear. The goal in the $x$BS1 problems is also to build a stack of height $x$, but all the blocks are not clear on the table initially in these problems—some blocks can be on top of each other. The initial state for `5BS1`, for example, initially puts block `I` on top of block `L`. Other $x$`BS1` problems have two or three pairs of blocks stacked initially, though there are no initial stacks of three or more blocks. Figure 5 shows initial and final states for two selected problems (complete specifications for the `nBS1` problems can be found elsewhere (Hanks & Weld, 1992; Kambhampati & Hendler, 1992).

---

9. See Section 9 for a discussion of the differences between the two systems.

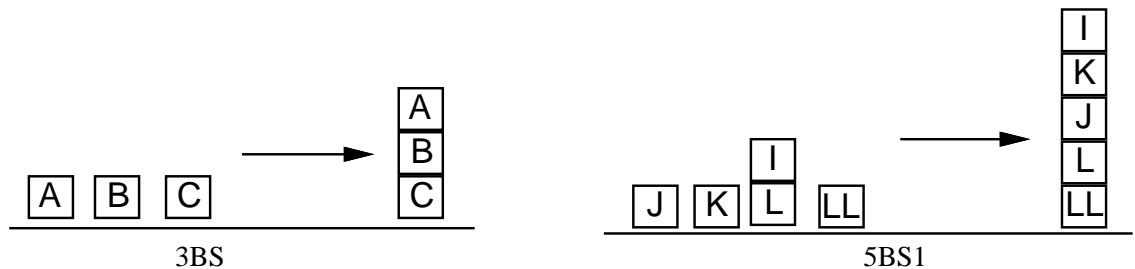

Figure 5: Two Blocksworld Problems

The PRIAR experiments involved comparing the planner's performance on a problem when the plan was generated from scratch with its performance when the solution to a smaller problem was used as a library plan. `4BS → 8BS` and `3BS → 5BS1` are two example experiments. For example, `3BS → 5BS1` involves comparing the time required to generate a plan for solving the `5BS1` from scratch with the time required for solving the `5BS1` problem starting with a solution for `3BS`.

Note that these experiments involve the adaptation process only—the problem of selecting an appropriate library plan was not considered.

### 8.1.1 Representation language

We tried to imitate PRIAR's representation language as closely as possible: both representations have two predicates, `ON` and `CLEARTOP`, and two primitive actions, `PUT-BLOCK-ON-BLOCK` and `PUT-BLOCK-ON-TABLE`.[10]

PRIAR uses a hierarchical representation, including non-primitive actions expressing concepts like "to get `A` on `B`, first generate a plan to clear `A`, then generate a plan to clear `B`, then execute the (primitive) `PUT-BLOCK-ON-BLOCK` action." SPA's representation consists only of descriptions for the two primitive actions. The closest analogue in SPA to hierarchical domain-specific knowledge is the notion of search-control information: application-supplied functions that determine which node in the graph of partial plans to consider next, what actions to introduce, in what order, how preconditions are to be achieved, and so on.

---

10. The domain theory presented in (Kambhampati & Hendler, 1992, Appendix B) also mentions pyramids and blocks, as well as various rules like nothing could be `ON` a pyramid. Since no pyramids figured in the experiments presented in (Kambhampati & Hendler, 1992, Section 7), we omitted them from our representation. PRIAR's representation also includes several domain axioms, e.g. one that defines `CLEARTOP` as the absence of one block `ON` another. SPA does not provide for domain axioms, so we incorporated that information into the action and problem definitions.

### 8.1.2 Control information

There is no obvious correspondence between PRIAR's hierarchical plan representation and SPA's control knowledge, so the question immediately arose as to what control information we should provide in running the experiments. SPA can exploit domain-dependent control information in three places:

1. to decide how to match objects in the (given) library plan against the objects in the input problem's initial and goal forms,

2. to decide which partial plan to consider next, and

3. to decide which part of the partial (incomplete) plan to work on next.

The first piece of domain-dependent control information involves how to fit the library plan to the new problem,[11] which involves choosing constants in the input problem to substitute for constants in the library plan. We adopted the same policy as did Kambhampati and Hendler: choose the substitution that maximizes the number of input goal forms that actually appear in the transformed library plan, and in the case of a tie choose the substitution that maximizes the number of initial conditions in the input problem that appear in the transformed library plan.

The problem is that finding the optimal mapping can be quite expensive: if the input problem mentions $n$ objects and the library problem mentions $k$ objects, finding the best mapping may involve examining all $\binom{n}{k}$ possibilities. The analysis in (Hanks & Weld, 1992) demonstrates the potential cost of mapping using the example of solving the 8BS problem using successively larger library plans. The complexity of computing the optimal mapping grows exponentially with the size of the library plan to the point where solving the 8BS problem using a solution to exactly the same problem as a library plan is actually more expensive than using a smaller library plan (even though it requires no adaptation at all). We note that this is similar to the *utility problem* addressed by Minton in the context of EBL (Minton, 1988). In subsequent experiments we used a heuristic, domain-dependent, linear-time mapping algorithm, described in (Hanks & Weld, 1992).

A control policy for the second decision requires shifting from breadth-first search to a best-first strategy. The longer paper discusses our ranking function in detail. To control decisions of the third sort (what flaw in the current plan to address next) we built a search-control heuristic that essentially implemented a policy of "build stacks from the bottom up." We acknowledge that the addition of domain specific heuristics complicates the comparison between SPA's performance and that of PRIAR, but we argue that this addition is "fair" because PRIAR used heuristic information itself. In PRIAR's case the domain specific knowledge took the form of a set of task-reduction schemata (Charniak & McDermott, 1984) rather than ranking functions, but both systems use heuristic control knowledge. Unfortunately, it is nearly impossible to assess the correspondence between the two forms of domain knowledge, but preliminary experiments, for example in (Barrett & Weld, 1994b), show that task-reduction schemata can provide planner speedup that is just as significant as that obtained by SPA's ranking functions.

---

11. Kambhampati and Hendler call this the *mapping* problem—it is well known in the literature, and is discussed in (Schank & Abelson, 1977), (Schank, 1982), and (Gentner, 1982) for example.

| Problem | Proc. time (msec) | | Speedup pctg. | |
|---|---|---|---|---|
| | SPA | PRIAR | SPA | PRIAR |
| 3BS→4BS1 | 1.7 | 2.4 | 59% | 40% |
| 3BS→5BS1 | 4.0 | 4.3 | 50% | 49% |
| 4BS→5BS1 | 2.9 | 3.2 | 64% | 62% |
| 4BS→6BS1 | 6.9 | 11.6 | 53% | 34% |
| 5BS→7BS1 | 11.2 | 11.1 | 58% | 71% |
| 4BS1→8BS1 | 18.6 | 22.2 | 55% | 72% |
| 4BS→8BS1 | 21.3 | 15.4 | 49% | 81% |
| 5BS→8BS1 | 19.2 | 10.1 | 54% | 87% |
| 6BS→9BS1 | 30.2 | 18.1 | 53% | 90% |
| 7BS→9BS1 | 24.9 | 11.4 | 61% | 94% |
| 4BS→10BS1 | 61.7 | 52.9 | 40% | 87% |
| 7BS→10BS1 | 40.7 | 23.4 | 61% | 94% |
| 8BS→10BS1 | 35.0 | 14.5 | 66% | 96% |
| 3BS→12BS1 | 133.2 | 77.1 | 18% | 96% |
| 5BS→12BS1 | 114.0 | 51.8 | 30% | 97% |
| 10BS→12BS1 | 53.1 | 21.2 | 67% | 99% |

Table 1: Comparative performance, SPA and PRIAR

## 8.2 Comparative results

The first three columns of Table 1 show how SPA's performance compares to PRIAR's in absolute terms.[12] We caution readers against using this information to draw any broad conclusions about the relative merits of the two approaches: the two programs were written in different languages, run on different machines, and neither was optimized to produce the best possible raw performance numbers.[13] Nonetheless we note that the absolute time numbers are comparable: SPA tended to work faster on smaller problems, PRIAR better on larger problems, but the data do not suggest that either program is clearly superior.

Kambhampati and Hendler assess PRIAR's performance relative to its own behavior in generating plans from scratch. This number, called the *savings percentage*, is defined to be as $\frac{s-r}{s}$, where $s$ is the time required to solve a problem, e.g. 12BS1, from scratch and $r$ is the time required to solve that same problem using a smaller library plan, e.g. one for 3BS. The fourth and fifth columns of Table 1 compare SPA and PRIAR on this metric.

The question therefore arises as to why PRIAR's speedup numbers are consistently so much larger in magnitude than SPA's, particularly on larger problems, even though absolute performance is not significantly better. The answer has to do with the systems' relative performance in planning from scratch. As Figure 6 demonstrates, PRIAR's performance degrades much faster than SPA's on generative tasks. We have no explanation for PRIAR's behavior, but its effect on the savings-percentage number is clear: these numbers are high because PRIAR's performance on generative tasks degrades much more quickly than does its behavior on refit tasks. Just to emphasize this relationship: for the 3BS→12BS1 problem PRIAR's processing time is 65% of SPA's. For 5BS→12BS1 it is 52% of SPA's. For 10BS→12BS1 the number is 43%, but for generating 12BS1 from scratch PRIAR runs about 12 times slower. This result points out that one must use extreme caution in evaluating any system based on these relative speedup figures, since they are actually measuring only the *relationship* between two separate components of a single system. It also points out that the problem of deciding when to generate plans from scratch instead of adapting them must take into account the effectiveness of the underlying generation mechanism.

---

12. All performance numbers for PRIAR appear in (Kambhampati & Hendler, 1992, Section 7).

13. See (Langley & Drummond, 1990) and (Hanks, Pollack, & Cohen, 1993) for a deeper discussion on the empirical evaluation of planning programs.

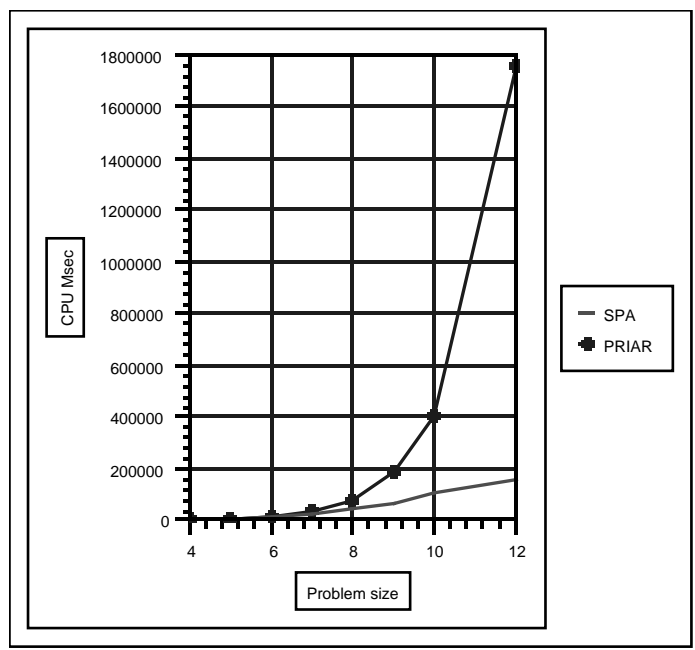

Figure 6: System performance for generative planning

## 8.3 Summary

Our two goals were to establish a systematic relationship between library use and problem-solving effort, and to compare our system's performance to that of the similar PRIAR. In the first case we note that on certain problems, most notably the nBS→mBS refits, there is a regular and systematic relationship between the fit between library and input problems (measured roughly by the difference between n and m) and the time required to solve the problem.[14] We should note, however, that the simple nature of the domain and the problems admits a particularly obvious measure of "degree of fit," so these results may not extend to less regular problem-solving scenarios. In the second case we demonstrated that the performance of the two systems was roughly comparable both in absolute terms and in terms of the relative value of refitting.

---

14. See (Hanks & Weld, 1992) for a deeper analysis, in which we develop a regression model that predicts the amount of time required to solve a problem based on the input and library plan sizes.

We must once again advise caution in interpreting these results. Although we believe they provide a preliminary validation of SPA's algorithm both in absolute terms and compared to PRIAR's hierarchical approach, the fact is the experiments were conducted in a very regular and simple problem domain, which allowed us to characterize the size of a problem or plan using a single number, and further allowed us to characterize the extent to which a library plan would be suitable for use in solving a larger input problem by comparing the numbers associated with the two plans.

Future work must therefore concentrate on two areas: the whole problem of how to retrieve a good plan from the library (which both SPA and PRIAR ignore), and the problem of assessing, in a realistic domain, the "degree of fit" between a library plan and an input problem. A similar analysis appears in (Koehler, 1994).

## 9. Related work

We have already mentioned the work on PRIAR (Kambhampati & Hendler, 1992) as close to our own, in particular its use of the generative planner to provide library plans and dependencies that can later be retracted. PRIAR and SPA also share the same STRIPS-like action representation. The main difference between the two approaches is the underlying planning algorithm: SPA uses a constraint-posting technique similar to Chapman's (1987) TWEAK, as modified by McAllester and Rosenblitt (1991), whereas PRIAR uses a variant of NONLIN (Tate, 1977), a hierarchical planner.

PRIAR's plan representation, and thus the algorithms that manipulate them, are more complicated that SPA's. There are three different types of validations (relationships between nodes in the plan graph), for example—filter condition, precondition, and phantom goal—as well as different "reduction levels" for the plan that represents a hierarchical decomposition of its structure, along with five different strategies for repairing validation failures. Contrast this representation with SPA's plan representation consisting of causal links and step-order constraints.

PRIAR's more complicated planning and validation structure makes it harder to evaluate the algorithm formally. Kambhampati and Hendler (1992, p. 39) prove a soundness result and argue informally for a property like completeness: "we claim that our framework covers all possible modifications for plans that are describable within the action representation described in this paper." It is not clear the exact relationship between this property and our completeness property.

The work on adaptation for case-based planning has mainly been concerned with finding good strategies for applying adaptations. In Section 7 we discussed CHEF (Hammond, 1990) in detail, analyzing it in terms of SPA's adaptation primitives. Since SPA uses the STRIPS representation and cannot represent simultaneous actions or actions with temporal extent, we were only able to consider ten of CHEF's seventeen repair strategies. However, we consider it interesting that nine of these transformations can be encoded simply as either one or two chained SPA primitives.

Section 2 also discussed PLEXUS (Alterman, 1988) and NoLimit (Veloso & Carbonell, 1993). Veloso (1992) also describes a mechanism by which case memory is extended during problem solving, including learning techniques for improving the similarity metric used in library retrieval. These issues have been completely ignored in our development of SPA, but it is possible that they could be added to our system.

Some case-based planning work, for example by Hammond (1990) and Alterman (1988), also addresses situations in which the planner's domain model is incomplete and/or incorrect. Both of these systems generate a plan using a process of retrieval and adaptation, then *execute* the plan. If execution fails (although the model incorrectly predicted that it would succeed), these systems try to learn the reasons why, and store the failure in memory so the system does not make the same mistake again. SPA sidesteps this challenging problem, since it addresses only the problem of ahead-of-time plan generation—not the problem of execution and error recovery. The XII planner (Golden, Etzioni, & Weld, 1994) uses a planning framework similar to SPA's, developing a representation and algorithm for generative planning in the presence of incomplete information; the XII planner still assumes what partial information it has is correct, however.

We mentioned in Section 2 that our goals in building the SPA system were somewhat different from most work in adaptive planning: our intent is that as a formal framework SPA can be used to analyze case-based planners to understand how they succeed in particular problem domains. As an implemented system we hope that SPA can be used to *build* effective problem solvers. The key is likely to be the addition of domain-dependent case-retrieval algorithms and heuristic control strategies.

## 10. Conclusions

We have presented the SPA algorithm, an approach to case-based planning based on the idea that the adaptation of previous planning episodes (library plans) is really a process of appropriately retracting old planning decisions and adding new steps, links and constraints in order to make the library plan skeleton solve the problem at hand.

The algorithm is simple, and has nice formal properties: soundness, completeness, and systematicity. It also makes clear the distinction between domain-independent algorithms and the application of domain-dependent control knowledge. As such it is an excellent vehicle for studying the problem of case-based planning in the abstract and for analyzing domain-dependent strategies for plan repair.

Our experimental results established a systematic relationship between computational effort required and the extent to which a library plan resembles the input problem, and also compared our system's performance to that of the similar PRIAR. The system's performance is encouraging, but we noted that the results should be interpreted within the context of the simple and regular problems in which they were conducted.

### 10.1 On the formal properties of algorithms

We should comment briefly on the implications of our algorithm's formal properties. Having properties like completeness and systematicity does not necessarily make an algorithm good, nor does the absence of these properties necessarily make an algorithm bad. The value of a framework for planning must ultimately be measured in its ability to solve interesting problems—to provide coverage of an interesting domain, to scale to problems of reasonable size, and so on. Soundness, completeness, and systematicity are neither necessary nor sufficient to build an effective planner.

However, the properties do help us to *understand* planning algorithms, however, which is equally important. What is it about CHEF that made it effective in its cooking domain? What is the essential difference between the PRIAR and the SPA frameworks? Formal analysis of an algorithm can provide insight into what makes it effective. We showed that CHEF's transformation strategies come at the cost of an incomplete algorithm, but understanding what parts of the search space they exclude can help us better understand how they are effective.

Formal properties can also act as an idealization of a desirable property that is more difficult to evaluate. Few would argue, for example, that systematicity is necessary for effective performance.[15] On the other hand, it is obviously important to make sure that a plan-adaptation algorithm does not cycle, and we can at least guarantee that a systematic algorithm will not cycle over partial plans.[16] So systematicity might be too strong a requirement for an algorithm, but at the same time it provides an end point in a spectrum.

## 10.2 Future work

Our work raises many questions that suggest avenues for future research:

- Although there are many hooks for domain-dependent information in our adaptation algorithm, we have not seriously explored the quality of the search-control interface. How convenient is it to specify heuristics to guide adaptation in a more realistic domain?

- Our analysis of transformational planning systems (Section 7) is preliminary. We hope to implement the approach described there and determine which of CHEF's transformational repairs provide the greatest computational benefit. It would also be interesting to perform the same type of analysis on GORDIUS (Simmons, 1988) or other transformational planners.

- The interplay between decisions made during the plan-retrieval process and the plan-adaptation process have not been well explored. We need to confront the issues faced by all case-based planners: what makes a good plan to retrieve, and what is the best way to fit that plan for the plan adapter? Our analysis (section 6.1) is an interesting start, but much is left to consider.

- One of the problems with the approach advocated in this paper is its dependence on the STRIPS action representation. It would be especially interesting to extend our ideas to a more expressive language (for example, something like ADL (Pednault, 1988) by adding retraction to UCPOP (Penberthy & Weld, 1992), or the language used by GORDIUS).

- The planning task is closely related to that of design (both are synthesis activities). We may be able to generalize our algorithm to address case-based design of lumped-parameter devices using ideas from system dynamics (Williams, 1990; Neville & Weld, 1992).

---

15. In fact empirical evidence, (Kambhampati, 1993), tends to suggest that systematic algorithms are actually less effective on common problems.

16. Though even a systematic plan-space planner can repeatedly generate plans that produce identical world states.

## Acknowledgements

This research was improved by discussions with Tony Barrett, Paul Beame, Denise Draper, Oren Etzioni, and Rao Kambhampati. Denise Draper cleaned up some of the code, infuriating us, but producing an improved system. David Madigan helped with the empirical analysis. Thanks also to Steve Minton, Alicen Smith, Ying Sun, and the anonymous reviewers, whose suggestions improved the presentation of this paper substantially. This work was funded in part by National Science Foundation Grants IRI-8902010, IRI-8957302, IRI-9008670, and IRI-9303461, by Office of Naval Research Grants 90-J-1904 and N00014-94-1-0060, and by a grant from the Xerox corporation.